\documentclass[11pt]{article}

\usepackage[preprint]{acl}

\usepackage{times}
\usepackage{latexsym}

\usepackage[T1]{fontenc}

\usepackage[utf8]{inputenc}

\usepackage{microtype}

\usepackage{inconsolata}

\usepackage{graphicx}

\usepackage{float}
\usepackage{multirow}

\usepackage{graphicx}
\usepackage{amsmath}
\usepackage{amssymb}
\usepackage{mathtools}

\usepackage{amsmath}
\usepackage{amsthm}

\usepackage{tcolorbox}
\tcbuselibrary{skins}
\usepackage{lipsum}
\usepackage{algorithm}
\usepackage{algpseudocode}
\usepackage{marvosym}
\usepackage{booktabs}

\theoremstyle{plain}
\newtheorem{theorem}{Theorem}[section]

\theoremstyle{definition}
\newtheorem{definition}[theorem]{Definition}

\theoremstyle{remark}
\newtheorem{remark}[theorem]{Remark}

\title{Teaching LLMs to Ask: Self-Querying Category-Theoretic Planning for Under-Specified Reasoning}

\author{Shuhui Qu \\
  Stanford University \\
  \texttt{shuhuiq@stanford.edu} \\
  }

\begin{document}
\maketitle
\begin{abstract}
Inference-time planning with large language models frequently breaks under partial observability: when task-critical preconditions are not specified at query time, models tend to hallucinate missing facts or produce plans that violate hard constraints.
We introduce \textbf{Self-Querying Bidirectional Categorical Planning (SQ-BCP)}, which explicitly represents precondition status (\texttt{Sat}/\texttt{Viol}/\texttt{Unk}) and resolves unknowns via (i) targeted self-queries to an oracle/user or (ii) \emph{bridging} hypotheses that establish the missing condition through an additional action.
SQ-BCP performs bidirectional search and invokes a pullback-based verifier as a categorical certificate of goal compatibility, while using distance-based scores only for ranking and pruning.
We prove that when the verifier succeeds and hard constraints pass deterministic checks, accepted plans are compatible with goal requirements; under bounded branching and finite resolution depth, SQ-BCP finds an accepting plan when one exists.
Across WikiHow and RecipeNLG tasks with withheld preconditions, SQ-BCP reduces resource-violation rates to \textbf{14.9\%} and \textbf{5.8\%} (vs.\ \textbf{26.0\%} and \textbf{15.7\%} for the best baseline), while maintaining competitive reference quality.
\end{abstract}

\section{Introduction}

Large language models (LLMs) can produce multi-step solutions via inference-time reasoning strategies such as chain-of-thought prompting \citep{wei2022chain}, tree-structured exploration \citep{yao2023tree}, and self-consistency sampling \citep{wang2022self}. Despite strong performance on curated benchmarks, these approaches typically treat the prompt as a complete specification of the task\citep{yao2022react}. In real user requests, however, crucial \emph{preconditions} are often missing (e.g., resource availability, hidden constraints, or unobserved environment facts)\citep{wei2025plangenllms}. In such settings, LLMs may fill gaps with plausible assumptions, yielding plans that are fluent but not executable\citep{zhang2025siren,ahn2022can}.

As a concrete example, consider ``Make a toy car from a wooden table.'' Whether a plan is feasible depends on facts not stated in the query: the table geometry, available tools, and permissible modifications. Standard prompting or search-based reasoning may still propose steps like ``cut the table leg into wheels'' without verifying (i) that a cutting tool exists, (ii) that the leg diameter is suitable, or (iii) that the operation does not violate the user's constraints. This illustrates a core failure mode of inference-time LLM planning under incomplete information: \emph{local coherence does not imply global feasibility} \citep{valmeekam2023planning, kamath2025enforcing, mullen2024lap}. 

Classical planning addresses feasibility by making preconditions explicit and checking them before applying actions \citep{fikes1971strips,aeronautiques1998pddl}. Handling incomplete information is also a central theme in conformant/contingent planning and POMDP formulations \citep{kaelbling1998planning,hoffmann2005contingent, palacios2009compiling}. However, directly transferring these formalisms to LLM planning is non-trivial: (i) LLMs typically \emph{generate} actions on the fly rather than selecting from a fixed operator library, and (ii) when applicability conditions are unknown, the system must \emph{actively resolve uncertainty} (e.g., via targeted questions or intermediate ``bridge'' steps) rather than silently assuming missing facts. Existing LLM-based information-seeking methods (e.g., question decomposition and querying) provide a useful ingredient \citep{press2023measuring}, but without an explicit precondition state and a hard verifier, unstructured querying can still miss critical constraints\citep{valmeekam2023planning,kambhampati2024can}.

We introduce \textbf{Self-Querying Bidirectional Categorical Planning (SQ-BCP)}, an inference-time planning framework that makes preconditions first-class and resolves uncertainty before committing to an action. SQ-BCP represents each candidate step as a hypothesis annotated with preconditions labeled \texttt{Sat}/\texttt{Viol}/\texttt{Unk}. When \texttt{Unk} preconditions are detected, SQ-BCP applies a deterministic refinement policy: it either asks a focused question to obtain the missing fact, or proposes a \emph{bridging} action that establishes the precondition (e.g., measuring, checking, or preparing) before proceeding. These locally refined hypotheses are integrated into a bidirectional search procedure that connects forward progress with backward-propagated goal requirements. Plan acceptance is \emph{not} based on a similarity heuristic; instead, SQ-BCP invokes a pullback-based categorical verifier as a black-box certificate of compatibility, together with deterministic hard-constraint checks. A task-dependent distance is used only for ranking and pruning during search, not as a correctness condition.

We evaluate SQ-BCP on two task sources that naturally exhibit missing preconditions: procedural instructions from WikiHow and recipe adaptation instances from RecipeNLG. We adopt a controlled $k$-reveal protocol that withholds annotated preconditions and requires methods to recover feasibility through querying or bridging. Across both sources, SQ-BCP substantially reduces resource-violation rates while maintaining competitive reference-similarity scores, demonstrating that explicit precondition tracking plus verification improves executability under partial observability.

Overall, the contributions are following:
\begin{enumerate}
    \item An inference-time planning framework that explicitly tracks reconditions with \texttt{Sat}/\texttt{Viol}/\texttt{Unk} semantics and resolves uncertainty via self-querying and bridging.
    \item A bidirectional integration that combines local precondition refinement with a pullback-based categorical verifier and deterministic hard-constraint checks, using heuristic distance only for ranking/pruning.
    \item An evaluation protocol on WikiHow and RecipeNLG with systematic precondition withholding, enabling controlled analysis of feasibility under partial observability.
\end{enumerate}
\section{Related Work}

\subsection{LLM Planning.}
LLMs can generate multi-step plans with inference-time prompting and search. Chain-of-Thought \citep{wei2022chain} and Self-Consistency \citep{wang2022self} improve deliberation via intermediate reasoning and sampling, while Tree-of-Thoughts (ToT) \citep{yao2023tree} explores branching solution candidates. ReAct \citep{yao2022react} further interleaves reasoning traces with actions and tool use. In embodied and robotic settings, approaches such as SayCan \citep{ahn2022can}, Inner Monologue \citep{huang2022inner}, and Code-as-Policies \citep{liang2022code} close the loop with environment feedback. More structured planners build graphs or explicit plan scaffolds from natural-language tasks \citep{huang2022language, sun2023adaplanner}. A common limitation is that feasibility is often treated implicitly: plans may be ranked by plausibility or reference similarity even when critical preconditions are missing or only partially observable. SQ-BCP targets this gap by explicitly tracking precondition status and enforcing hard-constraint checks and categorical verification before accepting a plan.

\subsection{Information Seeking and Question Generation.}
Generating questions to acquire missing information has a long history in both LLM reasoning and interactive IR. Self-Ask \citep{press2023measuring} decomposes problems into sub-questions but does not maintain an explicit, state-grounded representation of which action preconditions are satisfied versus unknown. Related work on clarification questions and conversational search focuses on resolving ambiguity or underspecification in user intents \citep{rao2018learning, aliannejadi2019asking, kim2024aligning}. Outside LLMs, active learning \citep{settles2009active, zhang2022active} formalizes information acquisition primarily during training, while information value theory \citep{howard2007information, krause2014submodular} characterize query utility given explicit probabilistic or reward models. SQ-BCP differs by operating \emph{at inference time} and tying question generation to \emph{action applicability}: queries are triggered by \texttt{Unk} preconditions and resolved into deterministic pass/fail outcomes for plan acceptance.

\subsection{Classical Planning and Preconditions.}
Classical planning formalizes actions with explicit preconditions and effects, as in STRIPS \citep{fikes1971strips} and PDDL \citep{aeronautiques1998pddl}, enabling efficient heuristic planners such as FF \citep{hoffmann2001ff} and Fast Downward \citep{helmert2006fast}. Partial observability is addressed through contingent planning with sensing actions \citep{hoffmann2005contingent}, conformant planning \citep{palacios2009compiling}, and POMDP formulations \citep{kaelbling1998planning}. These methods typically require symbolic states, hand-specified operator models, or reward structures, which are difficult to provide for open-ended natural language tasks. SQ-BCP borrows the spirit of explicit precondition checking, but instantiates it with coarse \texttt{Sat}/\texttt{Viol}/\texttt{Unk} labels and natural-language state representations, paired with inference-time information gathering (querying) and autonomous \emph{bridging} actions.

\subsection{Neuro-Symbolic and Compositional Reasoning.}
Neuro-symbolic approaches combine learning with symbolic structure, e.g., Neural Theorem Provers \citep{rocktaschel2016learning}, Neural Module Networks \citep{andreas2016neural}, and Logic Tensor Networks \citep{badreddine2022logic}, but they typically assume explicit predicates, modules, or differentiable logical operators. Separately, categorical methods provide principled abstractions for compositional structure and verification \citep{fong2023causal, shiebler2021categorical, tsukada2020computability,qu2025category}. SQ-BCP leverages categorical composition for plan construction and uses pullback-style verification as a compatibility certificate, avoiding the need for a fully symbolic domain model while still supporting compositional correctness checks.

\subsection{Verification and Constraint Satisfaction.}
Formal verification tools such as model checking \citep{clarke1997model} and SAT/SMT solving \citep{demoura2008smt} provide strong guarantees but require precise, formal specifications. In LLM pipelines, verification is often applied to final answers or intermediate reasoning (e.g., self-verification or program-aided checking) rather than to feasibility of action sequences \citep{weng2023largelanguagemodelsbetter, gao2023palprogramaidedlanguagemodels}. Classical planning metrics (e.g., makespan and plan distance/edit measures) characterize plan quality \citep{dwibedy2020onlineschedulingmakespanminimization} but do not directly address missing preconditions under natural-language task descriptions. SQ-BCP positions verification as an \emph{acceptance gate}: hard constraints are checked by exact predicates, and categorical verification certifies goal compatibility, while heuristic distances are used only for ranking/pruning.

\subsection{Planning Under Uncertainty and Interaction.}
Planning under uncertainty is classically studied via POMDP solvers and approximations \citep{Spaan_2005, NIPS2010_edfbe1af} and via active perception/next-best-view methods \citep{bajcsy1988active, bircher2016nbv}. In interactive learning, techniques such as DAgger \citep{ross2010dagger} and preference-based learning \citep{Sadigh2017ActivePL} incorporate feedback but typically rely on explicit policy or reward representations. Mixed-initiative planning and human--AI collaboration investigate coordination and explanation in interactive planning systems \citep{ferguson1998mixedinit, chakraborti2019explain}. SQ-BCP is complementary: it assumes no explicit reward model and operates zero-shot on natural-language tasks, using self-querying to reduce epistemic uncertainty about preconditions and bridging to establish missing conditions when possible.

SQ-BCP unifies three ingredients that are often treated separately: (i) explicit precondition uncertainty (\texttt{Sat}/\texttt{Viol}/\texttt{Unk}), (ii) state-grounded information seeking, and (iii) a compositional verification step for plan--goal compatibility. This combination is designed specifically for natural-language planning under partial observability, where feasibility depends on latent user/environment facts.

\section{Methodology}
\label{sec:method}
We formalize inference-time planning under partial observability by extending the categorical planning framework\citep{qu2025category} with explicit precondition uncertainty modeling and resolution. We present (i) hypothesis structure with labeled preconditions, (ii) deterministic refinement via self-querying and bridging, and (iii) pullback-based verification integrating into bidirectional categorical search.

\subsection{Problem Formulation}
\label{sec:problem}
We model task planning as a category \(\mathcal{T}\) whose objects are states and whose morphisms are valid operations. Each state
\[
w = (r, s, \ell, t) \in \mathcal{W}
\]
encapsulates resources \(r\), symbolic structure \(s\), logical predicates \(\ell\),
and temporal allocations \(t\). A morphism \(f:w_1\to w_2\) represents a valid state
transition that respects hard constraints over \((r,s,\ell,t)\).

\begin{definition}[Planning Problem]
\label{def:planning}
Given an initial state \(w_0=(r_0,s_0,\ell_0,t_0)\) and a goal specification
\(w^*=(r^*,s^*,\ell^*,t^*)\), find a sequence of morphisms in \(\mathcal{T}\):
\[
w_0 \xrightarrow{f_1} w_1 \xrightarrow{f_2} \cdots \xrightarrow{f_n} w_n
\]
such that each intermediate state \(w_i\) is feasible under the hard constraints and
the terminal state \(w_n\) is compatible with the goal in the sense certified by
pullback verification \citep{qu2025category}.
\end{definition}

\paragraph{Heuristic distance for ranking.}
We define a task-dependent distance to prioritize candidates:
\begin{equation}
\label{eq:distance}
\begin{split}
    D(w_1,w_2)=&\alpha_s d_s(s_1,s_2) +\alpha_r \|r_1-r_2\|_1 \\
  &+\alpha_\ell d_\ell(\ell_1,\ell_2)
+\alpha_t d_t(t_1,t_2).
\end{split}
\end{equation}
Importantly, \(D\) is used purely for \emph{ranking and pruning} during search.
Acceptance of a plan is determined by the verification procedure
(Section~\ref{sec:pullback_verify}), which checks hard constraints and categorical
compatibility independent of \(D\). In particular, if a chain fails hard checks or
pullback verification, it is rejected even when \(D(w_f,w^*)\) is small.

\paragraph{Partial observability and hypothesis chains.}
The goal specification may omit applicability facts required for candidate morphisms.
At inference time, the model produces hypotheses describing candidate morphisms along
with labeled preconditions and typed effects. Inference constructs a chain
\(H=h_1\cdot h_2\cdots h_n\), where each \(h_i\) is refined until its preconditions
are resolved. A chain is accepted only if (i) all preconditions are resolved,
(ii) all hard constraints are satisfied, and (iii) the verifier certifies categorical compatibility with the goal: the distance \(D(w_n,w^*)\!<\!\delta_{\text{accept}}\)

\subsection{State Space and Planning Category}
\label{sec:category}
The states are objects in \(\mathcal{T}\) and valid transitions are morphisms. 
SQ-BCP extends \citet{qu2025category} on the \emph{inference-time} procedure by introducing explicit precondition resolution:.

A morphism \(f:w\to w'\) applies typed effects:
\begin{equation}
    \begin{split}
        r' &= r + \Delta r,   s' = f(s)\\
        \ell' &= \ell \oplus \Delta \ell ,     t' = t + \Delta t.
    \end{split}
\end{equation}

\begin{figure}[t]
    \centering
    \includegraphics[width=\linewidth]{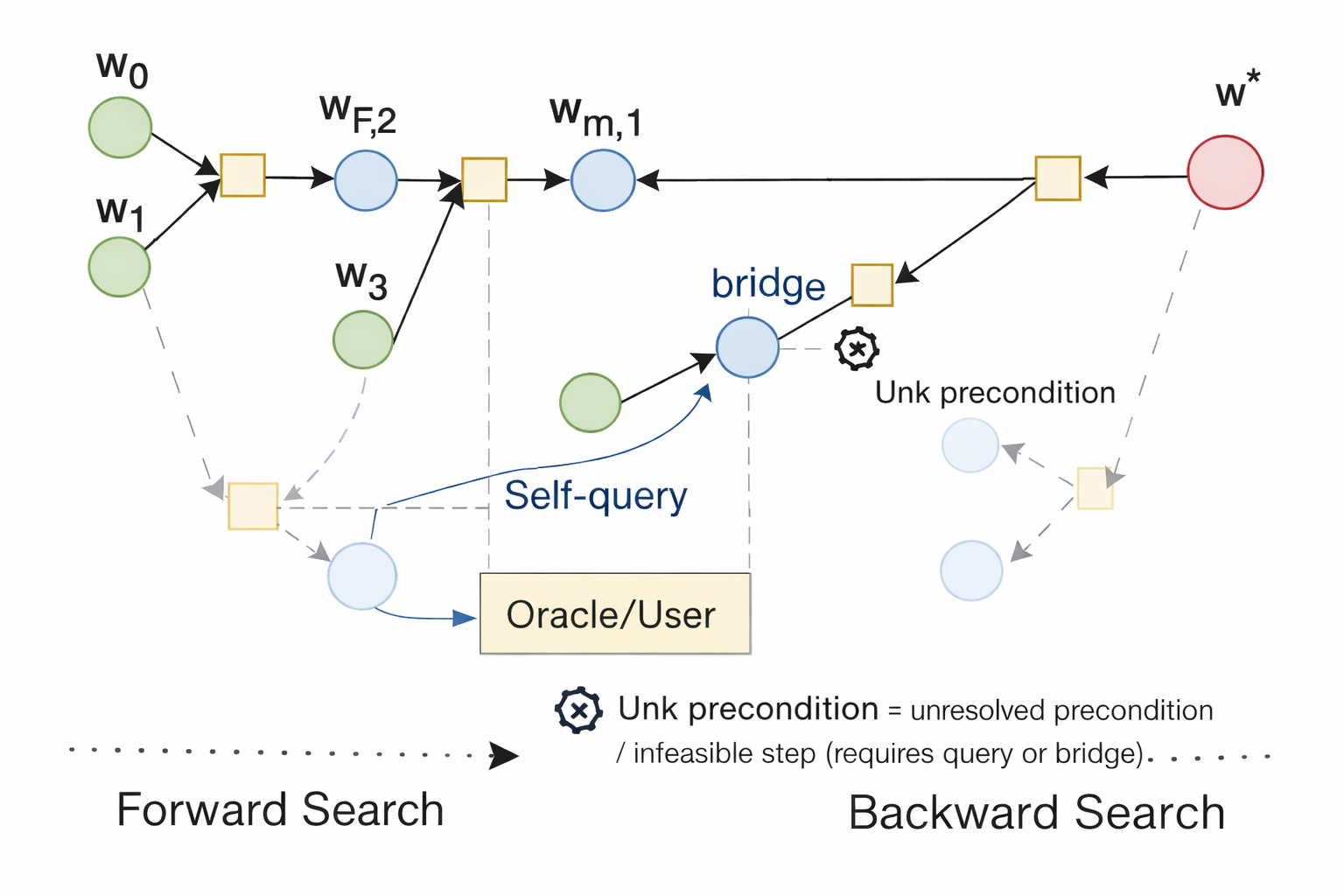}
    \caption{Bidirectional search from $w_0$ and $w^*$ with SQ-BCP self-querying and bridging hypotheses to resolve missing preconditions before categorical verification at meet points.}
    \label{fig:miss_precondition}
\end{figure}

\subsection{Hypothesis Structure}
\label{sec:hypothesis}

Given a state \(w_i\), the model generates a bounded set of candidate hypotheses:
\[
\mathcal{H}(w_i)=\{h_{i1},\ldots,h_{iK}\},
\]
each corresponding to a candidate morphism \(h:w_i\to w_{i+1}\).
A hypothesis is defined as \(h=(\textsc{action},\textsc{Pre},\textsc{Eff},\textsc{score})\), where \textsc{action} is a natural-language action, \textsc{Pre} is labeled preconditions, \textsc{Eff} is typed effects, and \textsc{score} is a scalar score in \([0,1]\).

\paragraph{Precondition labels.}
Each precondition is labeled \(\lambda_j\in\{\texttt{Sat},\texttt{Viol},\texttt{Unk}\}\).
Unknown preconditions are
\[
U(w_i,h)=\{p_j:(p_j,\texttt{Unk})\in \textsc{Pre}\}.
\]

\paragraph{Scoring for prioritization.}
We score a candidate hypothesis by its predicted progress:
\begin{equation}
\label{eq:score}
\textsc{score}(h)=\exp\!\left(-\frac{D(\textsc{Apply}(h,w_i),w^*)}{\tau}\right).
\end{equation}
This score is used to order expansions and apply conservative pruning thresholds;
it does not certify feasibility or correctness.

\subsection{Self-Querying and Bridging}
\label{sec:refine}

If \(U(w_i,h)\neq \emptyset\), SQ-BCP resolves uncertainty locally via:
\textbf{self-querying} (ask a targeted question for a missing fact) or
\textbf{bridging} (construct an auxiliary hypothesis \(h'\) whose effects establish
a missing precondition), producing a composed morphism \(h'\oplus h\). The score of the composition hypotheses is thus penalized as 
\[
\textsc{score}(h'\oplus h)=\min(\textsc{score}(h'),\textsc{score}(h))(1-\epsilon).
\]

\paragraph{Deterministic refinement policy.}
To keep cost accounting deterministic, we apply a fixed resolution order for each
unknown \(p\):
\begin{enumerate}
\item Attempt bridging up to \(T_{\text{bridge}}\) attempts.
\item If unresolved, issue query \(Q(p)\) once.
\item If the query is non-informative, mark \(p\) as unresolvable and discard \(h\).
\end{enumerate}
To prevent bridging loops, we track visited refinement signatures and terminate
bridging if a cycle is detected. Concretely, we maintain a hash set of signatures
\(\sigma=\langle w_i, p, \textsc{Pre\_labels}, \textsc{Eff\_summary}\rangle\); if a new bridge proposal yields a previously observed \(\sigma\), we halt bridging for \(p\) and fall back to querying. This makes refinement terminating under bounded attempts and avoids repeated bridge cycles.

\subsection{Bidirectional Search Integration}
\label{sec:bidirectional}
Our work follows the bidirectional search procedure due to it's efficiency. The bidirectional search pipeline maintains two AND-OR search graphs:
\begin{equation}
    \begin{split}
        G_F: &\text{ forward from } w_0,\\
        G_B: &\text{ backward from } w^*.
    \end{split}
\end{equation}

At each selected node, hypotheses are generated and refined \emph{locally} (querying/bridging) before being inserted as morphisms. Only hypotheses whose preconditions are resolved and that are not immediately rejected by hard constraints are eligible for graph insertion. This ensures that all morphisms in \(G_F\) and \(G_B\) are locally feasible at insertion time, though global compatibility still requires verification when forward and backward chains meet. Details can be found in \citep{qu2025category}

\subsection{Pullback-Based Verification}
\label{sec:pullback_verify}

To ensure compatibility, we perform a pullback-based verification procedure both constraint check and the distance screening.

\paragraph{Deterministic hard-constraint checks }
We apply deterministic constraint verification for each morphism to ensure:
(i) resource sufficiency in \(r_f\),
(ii) satisfaction of goal predicates in \(\ell_f\),
(iii) temporal feasibility in \(t_f\).
These checks are implemented as exact predicate evaluations (e.g., set containment for
resources, Boolean conjunction for logical predicates, explicit budget comparisons),
ensuring deterministic pass/fail outcomes independent of \(D\).

\paragraph{Distance screening (efficiency heuristic).}
To reduce computational cost by avoiding unnecessary verifier calls, we screen further candidates using distance function thresholds:
\begin{equation}
    \begin{split}
        \|r_f-r^*\|_1 < \delta_r,& d_s(s_f,s^*) < \delta_s,\\
        d_\ell(\ell_f,\ell^*) < \delta_\ell,&  d_t(t_f,t^*) < \delta_t.
    \end{split}
\end{equation}

\paragraph{Acceptance rule.}
A chain \(H\) is accepted when all three criteria been fulfilled:
(i) all preconditions across all states are resolved,
(ii) \(\textsc{HardCheck}(w_f,w^*)=\texttt{true}\),
(iii) \(\textsc{PullbackVerify}(w_f,w^*)=\texttt{true}\).

\subsection{Termination}
\label{sec:termination}
The search terminates under three conditions:(i) \textbf{Success:} an accepted chain (passes precondition resolution, hard checks,
and pullback verification). (ii)\textbf{Failure:} no frontier hypothesis remains above the pruning threshold, and (iii)\textbf{Timeout:} maximum expansion budget \(T_{\max}\) is reached.

\section{Theoretical Analysis}
\label{sec:theory}

We establish three core results: (1) precondition refinement terminates, (2) plans passing verification are correct, and (3) search finds solutions in bounded time under standard assumptions. 

\begin{theorem}[Refinement Terminates]
\label{thm:termination}
For any hypothesis \(h\) with finite unknowns \(U(w,h)\), the refinement procedure (Section~\ref{sec:refine}) terminates after at most \(|U(w,h)| \cdot (T_{\text{bridge}} + 1)\) steps, producing a hypothesis with all preconditions classified as \texttt{Sat} or \texttt{Viol}.
\end{theorem}

\begin{proof}
Each unknown \(p \in U(w,h)\) undergoes: (a) up to \(T_{\text{bridge}}\) bridging  attempts, then (b) one query. Cycle detection via signature hashing  \(\sigma = \langle w_i, p, \textsc{Pre\_labels},  \textsc{Eff\_summary}\rangle\)  prevents infinite bridging loops. Since \(U(w,h)\) is finite and each unknown is  processed with bounded attempts, refinement terminates in finite time.
\end{proof}

\begin{theorem}[Verification Certifies Correctness]
\label{thm:correctness}
Let chain \(H\) produce terminal state \(w_f\). If hard-constraint checks pass and \(\textsc{PullbackVerify}(w_f, w^*) = \texttt{true}\), then \(H\) is categorically compatible with the goal.
\end{theorem}

\begin{proof}
Hard checks verify resource sufficiency, logical predicate satisfaction, and temporal feasibility via exact predicate evaluation Section~\ref{sec:pullback_verify}).  Pullback verification then certifies compositional correctness via Theorems 3.1-3.3 in \citep{qu2025category}. Since our implementation applies both checks sequentially, passing verification implies the plan satisfies all goal requirements.
\end{proof}

\begin{remark}
Distance \(D(w_f, w^*) < \delta_{\text{accept}}\) is used only for screening and pruning; it does not imply correctness. A chain may pass the distance screen but fail verification (e.g., due to resource conflicts or structural incompatibilities not captured by the heuristic distance). Conversely, acceptance requires explicit verification independent of \(D\).
\end{remark}

\begin{theorem}[Bounded Search Complexity]
\label{thm:complexity}
Assume finite branching \(|\mathcal{H}(w)| \leq K\), bounded bridging depth \(R\), and at most \(|U_{\max}|\) unknowns per hypothesis. If a solution of depth \(d\) exists and no prefix on its path is pruned, then SQ-BCP finds it within \(T \leq K^d R^{|U_{\max}|}\) expansions.
\end{theorem}

\begin{proof}
At each of \(d\) states, explore at most \(K\) hypotheses. Each hypothesis has at most \(|U_{\max}|\) unknowns, each requiring at most \(R\) refinement attempts (bridging + query). Total expansions: \(K^d \cdot R^{|U_{\max}|}\). The no-pruning assumption (standard in heuristic search; cf. A* admissibility \citep{hart1968astar}) ensures the solution reachable.
\end{proof}

\begin{remark}
This bound is a worst-case guarantee assuming all unknowns require maximal refinement. In practice, (i) most hypotheses have few unknowns, (ii) bridging often succeeds quickly, and (iii) distance pruning eliminates poor branches early. The bound proves termination under bounded resources rather than actual runtime.
\end{remark}

These properties establish that SQ-BCP is a sound and terminating algorithm for inference-time planning under partial observability, with correctness guaranteed by explicit verification rather than heuristic distance.

\section{Experiments}
\label{sec:experiments}

We evaluate SQ-BCP on two benchmarks \textsc{RecipeNLG} and \textsc{WikiHow} designed to test precondition
reasoning under partial observability. 

\subsection{Benchmark Construction}
\label{sec:benchmark}

\paragraph{Domains and sources.}
We evaluate on two task sources that naturally expose latent preconditions.

\paragraph{Recipe Adaptation (RecipeNLG).}
We use the RecipeNLG dataset as the source of recipe-style planning tasks. Each instance specifies a goal dish along with available ingredients and a step-by-step procedure. We construct \textbf{recipe adaptation} tasks that
require ingredient substitution and cooking-method adjustment from foodKG \citep{foodkg}. Latent preconditions correspond to missing but necessary recipe facts (e.g., binding agent requirements, emulsification conditions, and hydration ratios). Tasks have 4-10 steps with 3-8 latent preconditions.

\paragraph{How-to tasks (WikiHow).}
We additionally adopt the WikiHow. We filter the dataset by keyword \textbf{``Things You'll Need''} and at least four procedure steps, yielding instances with explicit but incompletely specified resource and constraint requirements. Latent preconditions are instantiated primarily as withheld resource requirements (tools/materials) derived from \textbf{``Things You'll Need''}.

\paragraph{Instance structure.}
Each instance contains: (i) a minimal prompt with initial state and target, (ii) a fully specified prompt with complete resource and environment details, (iii) hard constraints, (iv) a set of  latent preconditions withheld from the underspecified variant, (substituent) (v) reference plan. 

\paragraph{$k$-reveal protocol.}
For an instance with $m$ latent preconditions $P=\{p_1,\ldots,p_m\}$, we create variants by revealing $k\in\{0,\ldots,m\}$ preconditions uniformly at random and injecting them into the prompt. The remaining $n=m-k$ must be discovered via querying or bridging. 

\subsection{Oracle Feedback Agent}
\label{sec:oracle}
There is an oracle agent that answers queries using ground-truth annotations. When an agent issues a query demonstrating \(Q\), the oracle agent returns the truthful answer and possible substitutions. All methods use identical oracle matching. For SQ-BCP, oracle calls are triggered when $|U(w,h)|>0$; for baselines, calls occur when they generate questions. To isolate structural gains from information access, we augment the
baselines that can receive identical oracle responses when they ask questions.

This oracle simulation models settings where users can answer factual questions about their constraints and resources (e.g., ``Do you have X tool?''), but cannot provide complete planning expertise.

\subsection{Baselines and Ablations}
\label{sec:baselines}

We compare against direct prompting, reasoning-augmented prompting, and search-augmented planners, all using GPT-4o unless noted:

\textbf{GPT-4o (Direct Prompting).} 
Prompted with raw task descriptions and request step-by-step plans, without additional reasoning instructions. 

\textbf{CoT (Chain-of-Thought).} 
Prompted with chain-of-thought. Explicit reasoning over resources, temporal requirements, and dependencies before producing a plan.

\textbf{ToT(Chain-of-Thought).} Prompted with chain-of-thought. Explicit reasoning over resources, temporal requirements, and dependencies before producing a plan.

\textbf{Thoughts-of-Search \citep{katz2024thought} } 
Structures LLM exploration as a guided search tree for improved reasoning depth.

\textbf{ReAct\citep{yao2022react}} 
Interleaves reasoning traces with environment interactions to refine planning decisions.

\textbf{Self-Ask} \citep{press2023measuring}: interleaves reasoning and
question generation.

All these methods share the same prompting template structure and have access to the oracle agent.

\subsection{Evaluation Metrics} \label{subsec:eval-metrics}
For WikiHow, we report: (1) \textbf{Rouge 1}: Overlap of the unigram between the result and label; (2) \textbf{Rouge 2}: Overlap of the bigram between the result and label; (3) \textbf{Constraint violation}: Percentage of solutions violating resource (i.e. use non-existing/wrong resources); For RecipeNLG: (4) \textbf{BLEU Score}; (5) \textbf{Constraint violations}: Percentage of solutions violating resource.

\section{Results}
\label{sec:results}

\subsection{Main Results}

Table~\ref{tab:main-results} reports results across task sources and
information levels, averaged over all $k\in[1,4]$ reveal settings.
Unless otherwise noted, all methods in Table~\ref{tab:main-results}
use the same backbone and configuration.

\begin{table*}[t]
\centering
\caption{Overall result (\%) across task sources, averaged over all
$k\in[1,4]$-reveal. Best results in \textbf{bold}, second best \underline{underlined}.}
\small
\begin{tabular}{l|ccc|cc}
\toprule
\multirow{2}{*}{\textbf{Method}} & \multicolumn{3}{c|}{\textbf{WikiHow}} & \multicolumn{2}{c}{\textbf{RecipeNLG}}  \\
 & ROUGE-1 ↑ & ROUGE-2 ↑ & Res Viol ↓  & BLEU ↑ & Res Viol ↓ \\
\midrule
Direct Prompt & 46.3 & 42.1 & 78.3 & 0.897 & 65.7  \\
CoT & 48.5 & 44.7 & 83.2 & 0.900 & 64.1 \\
ToT & 52.9 & 45.2 & 94.7 & 0.892 & 66.5 \\
ToS & 53.3 & 45.4& 82.6&0.898&64.2\\
ReAct & 55.8&47.4&76.9& 0.912&59.9\\
Self-Ask & \underline{56.1} & \underline{47.4} & 26.0 & \underline{0.913} & 15.7 \\
\textbf{SQ-BCP} & 52.7 & 45.9 & \textbf{14.9} & 0.907 & \textbf{5.8} \\
\bottomrule
\end{tabular}
\label{tab:main-results}
\end{table*}

\paragraph{Main results.}
Table~\ref{tab:main-results} shows a clear separation between reference-similarity
metrics (ROUGE/BLEU) and execution correctness (resource-violation rate). On both WikiHow and RecipeNLG, SQ-BCP achieves the lowest violation rate (\textbf{14.9}\% and \textbf{5.8}\%, respectively), substantially improving over the strongest baseline Self-Ask (26.0\% and 15.7\%). At the same time, SQ-BCP remains competitive on ROUGE/BLEU: while Self-Ask attains the best ROUGE/BLEU, SQ-BCP
matches or exceeds several planning baselines (e.g., CoT/ToT/ToS on RecipeNLG BLEU) and attains comparable ROUGE-2 on WikiHow. This pattern is consistent with SQ-BCP's design: candidates are ranked by heuristic distance and similarity, but \emph{accepted} only when hard constraints and verification succeed, which preferentially filters out plans that look close to the reference yet violate required resources.

\paragraph{Search and querying without structured verification can be misleading.}
Tree-style methods (ToT/ToS) and tool-augmented prompting (ReAct) obtain strong ROUGE on WikiHow, but still exhibit high violation rates (76.9--94.7\%),  ndicating that expanding more branches or interleaving actions does not reliably enforce feasibility when constraints are not explicitly represented. Self-Ask reduces violations substantially (26.0\% / 15.7\%) via question generation, but remains notably above SQ-BCP, suggesting that unstructured querying alone can still miss critical constraints. Overall, SQ-BCP shifts the Pareto frontier toward \emph{verified} executability: it trades a small drop in ROUGE/BLEU for a large reduction in constraint violations across both task sources.

\subsection{Ablation Study}

\begin{table}[t]
\centering
\small
\begin{tabular}{lcc}
\toprule
Method &   WikiHow Viol\%  & RecipeNLG Viol\% \\
\midrule
CoT (baseline)       & 83.2 & 64.1 \\
+ Pullback           & -- & -- \\
+ Self-Ask           & 26.0 & 15.7\\
SQ-BCP (full)        & 14.9 & 5.8 \\
\bottomrule
\end{tabular}
\caption{Ablation study. Each row adds a key component to a CoT-style backbone.}
\label{tab:ablation}
\end{table}

Table~\ref{tab:ablation} isolates how SQ-BCP’s components reduce constraint failures.Starting from a CoT baseline with high violation rates (83.2\% on WikiHow, 64.1\% on RecipeNLG), adding Self-Ask-style question generation yields the largest single drop (26.0\% / 15.7\%), confirming that information seeking is essential under partial observability. SQ-BCP further reduces violations to \textbf{14.9}\% / \textbf{5.8}\%, indicating that structured precondition semantics and verification provide additional benefit beyond asking questions---i.e., the remaining failures under Self-Ask are disproportionately due to missed or implicitly assumed hard constraints.

We do not report a standalone ``CoT + Pullback'' number because verification is not a drop-in improvement without a resolution mechanism for unknown preconditions: with no explicit \texttt{Unk} tracking and no querying/bridging policy, most plans either fail immediately at verification or degenerate into trivial rejection. This supports the intended interpretation of SQ-BCP: categorical verification is most effective when paired with explicit precondition modeling that can resolve missing applicability conditions before invoking the verifier.

\subsection{Model Scale and Architecture}

\begin{table}[t]
\centering
\small
\begin{tabular}{lcc}
\toprule
Model &  Wiki. Viol\%  & Recipe. Viol\% \\
\midrule
GPT-4o     & 14.9 & 5.8 \\
o4-mini  & 28.6 & 14.2 \\
Dpsk-R1-Dist.-Qwen-14B    & 10.5 & 4.3 \\
Qwen3-14B        & 11.4 & 5.2 \\
Claude 3.5  & 10.1 & 6.0 \\
\bottomrule
\end{tabular}
\caption{Cross-model performance.}
\label{tab:model-scale}
\end{table}

Table~\ref{tab:model-scale} evaluates SQ-BCP with different backbones under the same protocol. Across all five models, SQ-BCP maintains low violation rates, indicating that the gains from explicit precondition tracking and verification are not tied to a single model family. We observe the lowest violation rates with reasoning-oriented 14B-class models, suggesting that stronger intermediate reasoning and compliance with structured prompts improve precondition classification and downstream verification success. In contrast, the smaller o4-mini backbone shows higher violation rates, consistent with more frequent omissions or misclassifications of critical preconditions.

\subsection{Missing Preconditions}

\begin{figure}[t]
    \centering
    \includegraphics[width=0.85\linewidth]{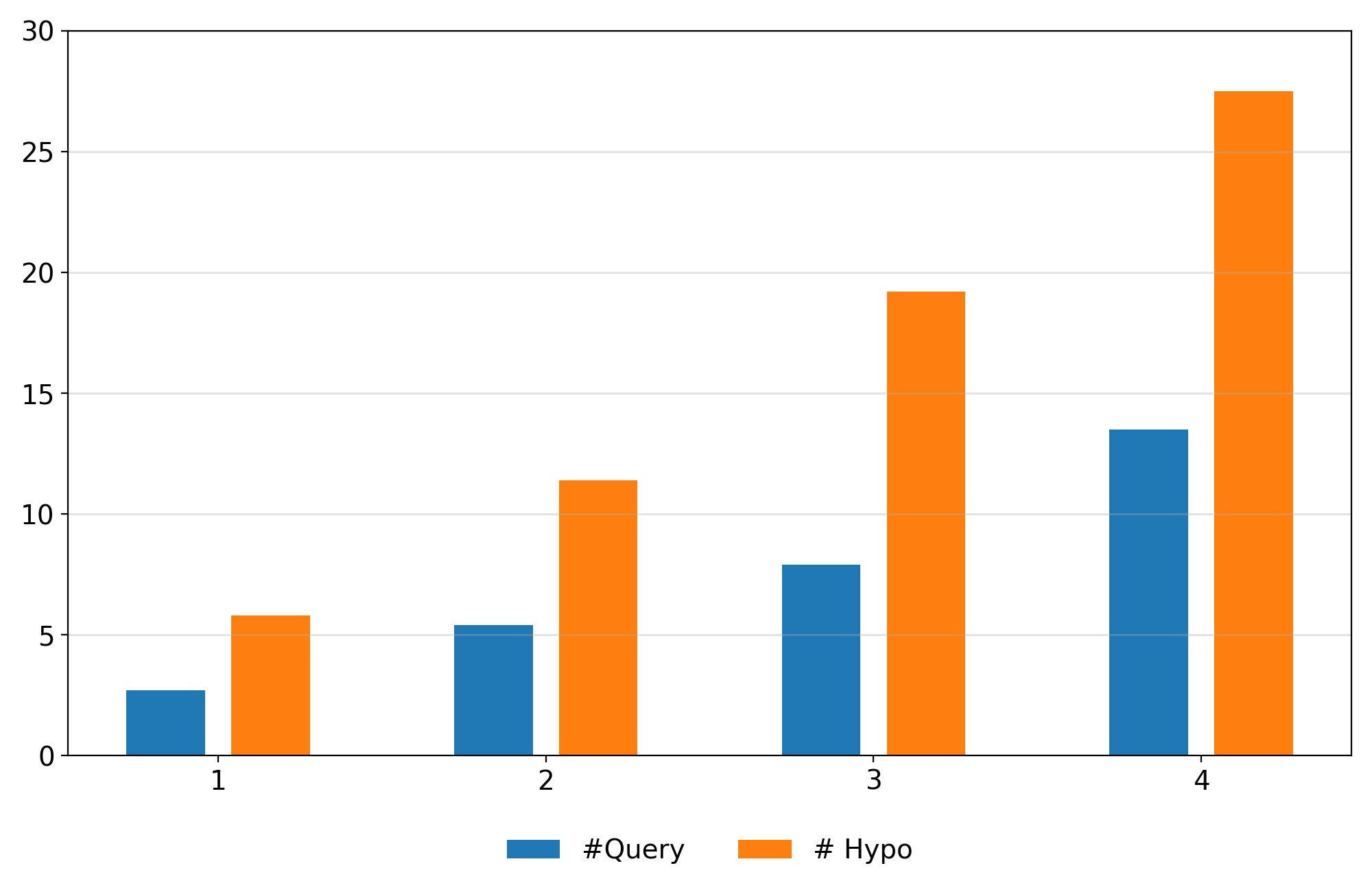}
    \caption{\textbf{Query and hypothesis counts vs. missing preconditions.} 
    As uncertainty increases, SQ-BCP generates more queries (blue) and explores     more hypotheses (green) to resolve unknowns.}
    \label{fig:res}
\end{figure}

Figure~\ref{fig:res} shows that SQ-BCP’s inference-time overhead grows smoothly as more preconditions are hidden: the number of oracle queries increases approximately linearly with the number of missing preconditions, consistent with our policy of issuing targeted questions only for \texttt{Unk} items after bounded local refinement.
In contrast, the number of hypotheses increases more steeply because each missing precondition can admit multiple candidate resolutions (e.g., alternative bridges),
which are enumerated locally before the verifier filters infeasible plans. Importantly, even in the hardest setting, the absolute counts remain moderate (about 13.5 queries and 27.3 hypotheses), suggesting that precondition resolution remains computationally tractable under the same executor.

\section{Conclusion}
\label{sec:conclusion}
We presented SQ-BCP, a planning framework for large language models under partial observability that explicitly represents precondition uncertainty and resolves it at inference time. SQ-BCP couples (i) state-grounded \texttt{Sat}/\texttt{Viol}/\texttt{Unk} precondition tracking, (ii) targeted self-querying and autonomous bridging actions to resolve unknowns, and (iii) a verification-based acceptance rule that enforces hard constraints and categorical plan--goal compatibility. Experiments on two task sources, WikiHow and RecipeNLG, show that SQ-BCP substantially reduces resource violations relative to strong prompting and search baselines while maintaining competitive plan quality under reference-similarity metrics. These results suggest that explicit precondition reasoning and verification are effective mechanisms for making LLM-generated plans more executable in settings where key facts are not available in the initial prompt.
\newpage
\section{Limitation}
\label{sec:limitations}
SQ-BCP is evaluated with a simulated oracle that answers factual queries about missing preconditions. While this approximates interactive settings where a user can confirm resource availability or constraint details, real users may provide incomplete, noisy, or inconsistent responses, and the interaction cost (latency, user burden) is not fully captured by our protocol. In addition, SQ-BCP depends on the backbone LLM to propose and label preconditions (\texttt{Sat}/\texttt{Viol}/\texttt{Unk}); mis-specified or misclassified preconditions can cause unnecessary queries, missed constraints, or premature rejection. 

Our verification operates over the discrete hard constraints implemented in the deterministic executor. Many real-world tasks involve continuous feasibility conditions (e.g., geometry, safety margins), stochastic effects, or underspecified success criteria that are difficult to express as exact predicates. Finally, SQ-BCP adds inference-time overhead from hypothesis generation, refinement, and verification, which may be costly for long-horizon tasks or latency-sensitive applications, and our empirical study focuses on procedural sources (WikiHow, RecipeNLG), so generalization to other domains remains to be validated.

\bibliography{custom}

\appendix
\onecolumn
\newpage

\section{Detailed Example}
\label{app:example}

\subsection{Complete Planning Trace}
\label{app:trace}

We illustrate SQ-BCP on a physical construction task with missing
preconditions, showing hypothesis generation, self-querying, bridging, and
pullback-based verification.

\paragraph{Task specification.}
Initial state $w_0$:
\begin{align*}
r_0 &= \{\texttt{wooden\_table}: 1, \texttt{saw}: 1, \texttt{ruler}: 1\},\\
s_0 &= \{\texttt{table\_legs}: 4,\; \texttt{leg\_shape}: \texttt{rectangular},\\
    &\quad\;\;\texttt{leg\_length}: \texttt{unknown},\;
           \texttt{leg\_diameter}: \texttt{unknown}\},\\
\ell_0 &= \{\texttt{workspace\_clear}: 1\},\\
t_0 &= (0, 7200)\quad\text{(2 hours available)}.
\end{align*}

Goal state $w^*$:
\begin{align*}
r^* &= \{\texttt{toy\_car}: 1\},\\
s^* &= \{\texttt{has\_wheels}: 1,\; \texttt{has\_body}: 1,\;
       \texttt{has\_axles}: 1,\; \texttt{assembled}: 1\},\\
\ell^* &= \{\texttt{functional}: 1,\; \texttt{safe\_for\_children}: 1\},\\
t^* &= (0, 7200).
\end{align*}

Goal constraints $C = \{c_r, c_s, c_\ell, c_t\}$:
\begin{itemize}
    \item $c_r$: final resource inventory must contain $\texttt{toy\_car}: 1$,
    \item $c_s$: structural requirements (wheels, body, axles, assembled),
    \item $c_\ell$: logical constraints (functional, safe),
    \item $c_t$: completion within time budget.
\end{itemize}

\paragraph{Distance and scoring parameters.}
We use:
\[
\alpha_r = 1.0,\quad \alpha_s = 2.0,\quad
\alpha_\ell = 1.5,\quad \alpha_t = 0.001.
\]
Temperature for scoring:
\[
\tau = 3.0,\quad\text{composition penalty } \varepsilon = 0.05.
\]
Score and distance thresholds are linked by
$\theta_{\min} = \exp(-\delta_{\text{accept}}/\tau)$.
For illustration, we set
\[
\theta_{\min} = 0.30,
\quad\Rightarrow\quad
\delta_{\text{accept}} \approx 3.61.
\]

\subsubsection*{Step 1: Initial Hypothesis Generation}

From state $w_0$, the LLM generates an initial hypothesis set
$H(w_0)$, including:

\paragraph{Hypothesis $h_1$.}
\begin{align*}
\text{action:}~ & \text{``cut table legs into wheels''}\\
\text{Pre:}~ & \{
   (\text{``legs are cylindrical''}, \text{Unk}),\\
   &\;\;(\text{``leg\_diameter suitable for wheels''}, \text{Unk}),\\
   &\;\;(\text{``saw available''}, \text{Sat})\},\\
\text{Eff:}~ & \Delta r = \{\texttt{wheels}: +4, \texttt{table\_legs}: -4\},\\
             & \Delta s = \{\texttt{has\_wheels}: 1\},\\
             & \Delta \ell = \emptyset,\quad
               \Delta t = +1800~\text{s}.
\end{align*}

Applying $h_1$ yields:
\[
w_1 = \mathrm{Apply}(h_1, w_0).
\]
We approximate the distance to the goal as:
\[
D(w_1,w^*) \approx 6.5,
\]
reflecting the fact that the agent has produced wheels but is still missing
body, axles, assembly, and safety guarantees.

With $\tau = 3.0$,
\[
\text{score}(h_1) = \exp\!\big(-D(w_1,w^*)/\tau\big)
                  = \exp(-6.5/3) \approx 0.12.
\]

\paragraph{Hypothesis $h_2$.}
\begin{align*}
\text{action:}~ & \text{``use table top as car body''}\\
\text{Pre:}~ & \{
   (\text{``table\_top detachable''}, \text{Unk}),\\
   &\;\;(\text{``table\_top size suitable''}, \text{Unk})\},\\
\text{Eff:}~ & \Delta r = \{\texttt{car\_body}: +1, \texttt{table}: -1\},\\
             & \Delta s = \{\texttt{has\_body}: 1\},\\
             & \Delta \ell = \emptyset,\quad
               \Delta t = +900.
\end{align*}
We approximate $D(\mathrm{Apply}(h_2,w_0), w^*) \approx 6.8$, yielding
\[
\text{score}(h_2) = \exp(-6.8/3) \approx 0.11.
\]

\paragraph{Hypothesis $h_3$.}
\begin{align*}
\text{action:}~ & \text{``purchase pre-made toy car kit''}\\
\text{Pre:}~ & \{
   (\text{``budget available''}, \text{Unk}),\\
   &\;\;(\text{``store nearby''}, \text{Unk})\},\\
\text{Eff:}~ & \Delta r = \{\texttt{toy\_car}: +1\},\\
             & \Delta s = \{\texttt{has\_wheels}:1,\texttt{has\_body}:1,\\
             &\qquad\;\;\texttt{has\_axles}:1,\texttt{assembled}:1\},\\
             & \Delta \ell = \{\texttt{functional}:1,
                               \texttt{safe\_for\_children}:1\},\\
             & \Delta t = +3600.
\end{align*}
Here $D(\mathrm{Apply}(h_3,w_0),w^*) \approx 0.5$ and:
\[
\text{score}(h_3) = \exp(-0.5/3) \approx 0.85.
\]

At this point, all hypotheses have unknown preconditions:
\begin{align*}
U(w_0,h_1) &= \{\text{``legs are cylindrical''},\;
               \text{``leg\_diameter suitable''}\},\\
U(w_0,h_2) &= \{\text{``table\_top detachable''},\;
               \text{``table\_top size suitable''}\},\\
U(w_0,h_3) &= \{\text{``budget available''},\;
               \text{``store nearby''}\}.
\end{align*}
Hence $\Phi(w_0,h_i)=0$ for all $i$, and SQ-BCP must resolve
preconditions. We start with the highest-scoring hypothesis $h_3$.

\subsubsection*{Step 2: Self-Querying for $h_3$}

SQ-BCP generates queries:
\begin{align*}
Q_1 &: \text{``What is your budget for this project?''}\\
Q_2 &: \text{``Is there a toy store within reasonable distance?''}
\end{align*}
User responses:
\begin{align*}
A_1 &: \text{No budget -- must use existing materials.}\\
A_2 &: \text{Not applicable.}
\end{align*}

We update preconditions of $h_3$:
\[
\text{Pre}(h_3') =
\{(\text{``budget available''}, \text{Viol}),\;
  (\text{``store nearby''}, \text{Unk})\}.
\]
Because one precondition is violated, $h_3$ becomes ineligible and is
removed from $H^*(w_0)$.

\subsubsection*{Step 3: Self-Querying for $h_1$}

Next we resolve $h_1$. Queries:
\begin{align*}
Q_1 &: \text{``What is the cross-sectional shape of the table legs?''}\\
Q_2 &: \text{``What is the diameter of the table legs?''}
\end{align*}
User responses:
\begin{align*}
A_1 &: \text{Rectangular cross-section, approximately 3cm $\times$ 3cm.}\\
A_2 &: \text{Same as above -- rectangular, not cylindrical.}
\end{align*}
Update:
\[
\text{Pre}(h_1') =
\{(\text{``legs are cylindrical''}, \text{Viol}),\;
  (\text{``leg\_diameter suitable''}, \text{Viol}),\;
  (\text{``saw available''}, \text{Sat})\}.
\]
Thus $h_1'$ is ineligible and removed.

\subsubsection*{Step 4: Bridging for $h_1$ (Alternative Path)}

Instead of querying, SQ-BCP may attempt bridging. For
$p_1=\text{``legs are cylindrical''}$, a bridging hypothesis $h_1'$:
\begin{align*}
\text{action:}~ &
\text{``reshape table legs into cylinders using lathe''}\\
\text{Pre:}~ &
\{(\text{``lathe available''}, \text{Unk}),\;
  (\text{``table\_legs removable''}, \text{Sat})\},\\
\text{Eff:}~ &
\Delta s = \{\texttt{leg\_shape}: \texttt{cylindrical},
              \texttt{leg\_diameter}:25\text{mm}\},\\
&\Delta r = \emptyset,\quad \Delta \ell = \emptyset,\quad
\Delta t = +1200.
\end{align*}
The system queries:
\[
Q: \text{``Is a lathe available?''}
\]
User answer:
\[
A: \text{No, only hand tools.}
\]
Hence $\text{Pre}(h_1')$ includes a violated precondition and $h_1'$ is
discarded.

An alternative bridging hypothesis $h_1''$:
\begin{align*}
\text{action:}~ &
\text{``sand/carve table legs into approximate cylinders''}\\
\text{Pre:}~ &
\{(\text{``sandpaper available''}, \text{Sat}),\;
  (\text{``manual effort acceptable''}, \text{Sat})\},\\
\text{Eff:}~ &
\Delta s =
\{\texttt{leg\_shape}: \texttt{roughly\_cylindrical},
  \texttt{leg\_diameter}: 25\text{mm}\},\\
&\Delta r = \emptyset,\quad \Delta \ell = \emptyset,\quad
\Delta t = +2400.
\end{align*}
We approximate $D(\mathrm{Apply}(h_1'',w_0),w^*) \approx 7.8$, hence
\[
\text{score}(h_1'') = \exp(-7.8/3) \approx 0.08.
\]

Composed hypothesis $h_{1,\text{comp}} = h_1'' \oplus h_1$:
\begin{align*}
\text{action:}~ &
\text{``First sand legs into cylinders, then cut into wheels''},\\
\text{Pre:}~ &
\{(\text{``sandpaper available''},\text{Sat}),\;
  (\text{``manual effort acceptable''},\text{Sat}),\;
  (\text{``leg\_diameter suitable''},\text{Sat}),\;
  (\text{``saw available''},\text{Sat})\},\\
\text{Eff:}~ &
\Delta r = \{\texttt{wheels}: +4, \texttt{table\_legs}: -4\},\\
&\Delta s =
\{\texttt{has\_wheels}: 1, \texttt{leg\_shape}: \texttt{cylindrical}\},\\
&\Delta \ell = \emptyset,\quad
\Delta t = 2400 + 1800 = 4200.
\end{align*}

Score:
\[
\text{score}(h_{1,\text{comp}})
= \min(\text{score}(h_1''), \text{score}(h_1))(1-\varepsilon)
= \min(0.08, 0.12)\cdot 0.95 = 0.076.
\]
All preconditions are now satisfied, so
$\Phi(w_0,h_{1,\text{comp}})=1$.

\subsubsection*{Step 5: Apply Composed Hypothesis}

Applying $h_{1,\text{comp}}$:
\[
w_1 =
(r_1,s_1,\ell_1,t_1)
=
(\{\texttt{wooden\_table}:1,\texttt{saw}:1,\texttt{ruler}:1,\texttt{wheels}:4\},\;
 \ldots,\; (0,4200)).
\]
In particular:
\[
s_1:\ \texttt{has\_wheels}=1,\quad
\ell_1:\ \texttt{workspace\_clear}=1.
\]

At $w_1$, the LLM proposes new hypotheses $H(w_1)$. We focus on $h_4$:

\paragraph{Hypothesis $h_4$.}
\begin{align*}
\text{action:}~ &
\text{``cut table top into car body base''}\\
\text{Pre:}~ &
\{(\text{``saw available''}, \text{Sat}),\;
  (\text{``table\_top dimensions suitable''}, \text{Unk})\},\\
\text{Eff:}~ &
\Delta r = \{\texttt{car\_body}: +1, \texttt{table}: -1\},\\
&\Delta s = \{\texttt{has\_body}: 1\},\quad
\Delta \ell = \emptyset,\quad
\Delta t = +1200.
\end{align*}

We approximate
$D(\mathrm{Apply}(h_4,w_1),w^*) \approx 4.5$, giving
$\text{score}(h_4) \approx \exp(-4.5/3) \approx 0.22$.

A query resolves the unknown:
\[
Q: \text{``What are the dimensions of the table top?''}
\]
\[
A: \text{``60cm $\times$ 40cm, too large for a toy car.''}
\]
Thus $(\text{``table\_top dimensions suitable''},\text{Viol})$ and
$h_4$ becomes ineligible.

\subsubsection*{Step 6: Bridging for $h_4$ and Move to $w_2$}

We introduce a bridging hypothesis $h_4'$:
\begin{align*}
\text{action:}~ &
\text{``cut a smaller section from table top''}\\
\text{Pre:}~ &
\{(\text{``saw available''}, \text{Sat})\},\\
\text{Eff:}~ &
\Delta r = \{\texttt{wood\_piece}: +1\},\\
&\Delta s = \{\texttt{suitable\_body\_piece}: 1\},\\
&\Delta \ell = \emptyset,\quad \Delta t = +600.
\end{align*}
with $\text{score}(h_4') \approx \exp(-7.2/3) \approx 0.09$.

Then a modified hypothesis $h_{4,\text{body}}$:
\begin{align*}
\text{action:}~ &
\text{``shape wood piece into car body''}\\
\text{Pre:}~ &
\{(\text{``suitable\_body\_piece''}, \text{Sat}),\;
  (\text{``saw available''}, \text{Sat})\},\\
\text{Eff:}~ &
\Delta r = \{\texttt{car\_body}: +1, \texttt{wood\_piece}: -1\},\\
&\Delta s = \{\texttt{has\_body}: 1\},\quad
\Delta \ell = \emptyset,\quad
\Delta t = +900.
\end{align*}

The composed hypothesis
$h_{4,\text{comp}} = h_4' \oplus h_{4,\text{body}}$ has:
\[
\text{score}(h_{4,\text{comp}})
= \min(0.09, 0.22)\cdot 0.95 = 0.086,
\quad
\Phi(w_1,h_{4,\text{comp}})=1.
\]

Applying $h_{4,\text{comp}}$ yields $w_2$:
\[
r_2 = \{\texttt{saw}:1,\texttt{ruler}:1,\texttt{wheels}:4,\texttt{car\_body}:1\},
\]
\[
s_2:~ \texttt{has\_wheels}=1,\;\texttt{has\_body}=1,\;\texttt{assembled}=0,\;
      \texttt{has\_axles}=0,
\]
\[
t_2 = (0,5700).
\]

\subsubsection*{Step 7: Final Step to the Goal}

At $w_2$, a natural hypothesis $h_5$:
\begin{align*}
\text{action:}~ &
\text{``create axles from leg remnants and attach wheels to body''}\\
\text{Pre:}~ &
\{(\text{``remnants available''}, \text{Sat}),\;
  (\text{``wheels compatible with body''}, \text{Sat})\},\\
\text{Eff:}~ &
\Delta r = \{\texttt{toy\_car}: +1,\texttt{wheels}: -4,\texttt{car\_body}: -1\},\\
&\Delta s = \{\texttt{has\_axles}:1,\texttt{assembled}:1\},\\
&\Delta \ell = \{\texttt{functional}:1,\texttt{safe\_for\_children}:1\},\\
&\Delta t = +1200.
\end{align*}

Applying $h_5$ yields $w_5$:
\begin{align*}
r_5 &= \{\texttt{saw}:1,\;\texttt{ruler}:1,\;\texttt{toy\_car}:1\},\\
s_5 &= \{\texttt{has\_wheels}:1,\;\texttt{has\_body}:1,\;
       \texttt{has\_axles}:1,\;\texttt{assembled}:1\},\\
\ell_5 &= \{\texttt{workspace\_clear}:1,\;\texttt{functional}:1,\;
          \texttt{safe\_for\_children}:1\},\\
t_5 &= (0, 6900).
\end{align*}

We approximate:
\[
D(w_5,w^*) \approx 1.1
\quad(< \delta_{\text{accept}} \approx 3.61),
\]
and
\[
\text{score}(h_5) = \exp(-1.1/3) \approx 0.69.
\]

\subsubsection*{Step 8: Pullback Verification}

The hypothesis chain is:
\[
H = (h_1'' \oplus h_1) \cdot (h_4' \oplus h_{4,\text{body}}) \cdot h_5.
\]

All preconditions at each step are resolved:
\[
U(w_0,h_1''\oplus h_1) = \emptyset,\quad
U(w_1,h_4'\oplus h_{4,\text{body}}) = \emptyset,\quad
U(w_2,h_5) = \emptyset.
\]

Component-wise distances:
\begin{itemize}
    \item $c_r$: $r_5$ vs.\ $r^*$ differ only by extra tools, which
          do not violate resource constraints; $\|r_5-r^*\|$ is small.
    \item $c_s$: $s_5 = s^*$, so $d_s(s_5,s^*)=0$.
    \item $c_\ell$: $\ell_5$ contains $\ell^*$ plus workspace info, so
          $d_\ell(\ell_5,\ell^*) \approx 0$.
    \item $c_t$: $t_5$ and $t^*$ differ by $300$ seconds;
          $d_t(t_5,t^*)=300 < \delta_t$.
\end{itemize}

A pullback
\[
P = w_5 \times_C w^*
\]
thus exists and is valid in $\mathcal{T}$. Since
$D(w_5,w^*) < \delta_{\text{accept}}$ and
$\exp(-D(w_5,w^*)/\tau) \approx 0.69 \ge \theta_{\min}$, SQ-BCP accepts $H$
as a valid plan.

\subsection{Key Insights from the Example}
\label{app:insights}

This example highlights several qualitative properties of SQ-BCP:

\begin{itemize}
    \item \textbf{Self-querying prevents infeasible paths.}
    Querying for $h_3$ quickly reveals that the budget constraint is violated,
    avoiding search in an unreachable branch.

    \item \textbf{Bridging enables blocked hypotheses.}
    Hypothesis $h_1$ is initially blocked by unknown shape/diameter
    preconditions. Bridging via $h_1''$ constructs a feasible refinement
    $h_{1,\text{comp}}$ that establishes these preconditions and makes the
    original high-level operation applicable.

    \item \textbf{Score and composition.}
    Composed hypotheses inherit the weakest component score (up to a penalty),
    focusing the planner on chains where each step contributes meaningfully.
    In practice, acceptance is decided by the \emph{final} state distance and
    its induced score, consistent with the distance--score coupling
    $\theta_{\min} = \exp(-\delta_{\text{accept}}/\tau)$.

    \item \textbf{Query vs.\ bridge trade-off.}
    Queries are cheap but rely on external answers; bridging is autonomous but
    increases plan complexity and time cost. SQ-BCP naturally mixes both:
    querying for $h_3$ (fast rejection) and bridging for $h_1$ (successful
    refinement).

    \item \textbf{Pullback as a global guarantee.}
    Pullback-based verification ensures that, regardless of local scoring,
    the final plan satisfies the global resource, structural, logical, and
    temporal constraints. Distance $D$ is used to prioritize candidates and
    screen for likely pullbacks; the pullback itself provides the formal
    correctness guarantee.
\end{itemize}

\section{Full Experimental Results}
\label{app:full-results}

This appendix includes the expanded results omitted from the main paper
due to space constraints.

\section{Implementation Details}
\label{sec:implementation}

\paragraph{Distance function (ranking and pruning, not correctness).}
We use Eq.~\eqref{eq:distance} with weights $\alpha_r=1.0$, $\alpha_s=2.0$,
$\alpha_\ell=1.5$, $\alpha_t=0.001$. Component distances:
$d_r$ = L1 norm on resource counts;
$d_\ell$ = Hamming distance on logical bit-vectors;
$d_t$ = absolute difference in completion times;
$d_s$ = cosine distance on sentence embeddings.
$D$ is used purely for \emph{ranking and pruning} during search.
Acceptance of a plan is determined by the verification procedure
(Section~\ref{sec:pullback_verify}), which checks hard constraints and categorical compatibility independent of $D$.

\paragraph{Screening and scoring parameters.}
Screening thresholds: $\delta_{\text{accept}}=3.5$, $\delta_r=1.5$, $\delta_s=0.7$,
$\delta_\ell=2.0$, $\delta_t=3600$. Scoring parameters: $\tau=3.0$, $\epsilon=0.05$,
$\theta_{\min}=\exp(-\delta_{\text{accept}}/\tau)\approx 0.31$.

\paragraph{Hard constraint verification.}
Before pullback verification, we apply deterministic hard-constraint checks:
(i) resource sufficiency ($r^*\subseteq r_f$ via set containment),
(ii) logical satisfaction ($\ell^*\subseteq \ell_f$ via Boolean conjunction),
(iii) temporal feasibility ($t_f\leq t_{\text{budget}}$ via threshold comparison).
These checks are implemented as exact predicate evaluations rather than soft
similarity measures, ensuring deterministic pass/fail outcomes independent of $D$.

\end{document}